\documentclass{article}
\usepackage{spconf,amsmath,graphicx,hyperref}
\usepackage{multirow}
\usepackage{subcaption}
\usepackage{array,etoolbox}
\usepackage{booktabs}
\usepackage{bm}
\usepackage{pgfplots}
\pgfplotsset{compat=1.18}
\usepackage{pgfplotstable}
\usepackage{filecontents}
\usepackage{makecell}

\title{BEST-RQ-Based Self-Supervised Learning for Whisper domain adaptation}
\name{Raphaël Bagat, Irina Illina, Emmanuel Vincent\thanks{This work was funded by the DeepMAUVES project supported by DGA of french MoD and CNRS, and granted access to the HPC resources of IDRIS under the allocation 2024-AD011015024 made by GENCI.}}
\address{Université de Lorraine, CNRS, Inria, LORIA, F-54000 Nancy, France}
\begin{document}
\ninept

© 20XX IEEE. Personal use of this material is permitted. Permission from IEEE must be obtained for all other uses, in any current or future media, including reprinting/republishing this material for advertising or promotional purposes, creating new collective works, for resale or redistribution to servers or lists, or reuse of any copyrighted component of this work in other works.

\newpage
\maketitle
\begin{abstract} % about 100 - 150 words
Automatic Speech Recognition (ASR) systems, despite large multilingual training, struggle in low-resource scenarios where labeled data is scarce. 
We propose BEARD (BEST-RQ Encoder Adaptation with Re-training and Distillation), a novel framework designed to adapt Whisper’s encoder with unlabeled data.
Unlike traditional self-supervised learning methods, BEARD uniquely combines a BEST-RQ objective with knowledge distillation from a frozen teacher encoder, ensuring the encoder's complementarity with the pre-trained decoder.
Our experiments focus on the ATCO2 corpus from the challenging Air Traffic Control (ATC) communications domain, characterized by non-native speech, noise, and specialized phraseology. 
Using about 5,000 hours of untranscribed speech for BEARD and 2 hours of transcribed speech for fine-tuning, the proposed approach significantly outperforms previous baseline and fine-tuned model, achieving a relative improvement of 12\% compared to the fine-tuned model. 
To the best of our knowledge, this is the first work to use a self-supervised learning objective for domain adaptation of Whisper.
\end{abstract}
\begin{keywords}
Automatic speech recognition, Whisper, self-supervised learning, domain adaptation
\end{keywords}
\section{Introduction}
\label{sec:intro}

Automatic speech recognition (ASR) has reached near-human accuracy in many domains \cite{xiong2017toward}.
The arrival of large‐scale end-to-end models made these models easier to use out-of-the-box.
However, despite being trained on massive multilingual datasets, these models still struggle with out-of-domain scenarios, like out-of-vocabulary words, spontaneous speech, noisy speech, etc. 
%that may not have been sufficiently represented during training (CHANGER REF). 
To adapt models to these new domains, supervised, self-supervised or self- trainings can be explored. 
Supervised training needs transcribed audio, which is costly and time-consuming to obtain.

A way to improve a speech recognition system to a new domain with little transcribed speech and a larger amount of untranscribed audio, is self-training \cite{zavaliagkos1998utilizing}.
A teacher model is first trained using the transcribed audio, and then used to create \textit{pseudo-labels} for the untranscribed speech.
A student model is then trained using the pseudo-labels in a supervised way.
Though this method has shown to be successful \cite{kahn2020self}, pseudo-labeling large amounts of data is computationally expensive and sensitive to the teacher's accuracy.

Self-supervised learning (SSL) methods were introduced to perform model training in the absence of annotations.
SSL methods enable leveraging large amounts of unlabeled data to learn informative representations through representation learning.
%aim to encode the speech signal into informative representations through representation learning.
%For Transformer-based models, a decoder can then be added on top of the SSL encoder and trained in a supervised manner, which has been shown to improve speech recognition performance \cite{chung2021w2v}.
A decoder can then be added on top of the SSL encoder and trained in a supervised manner for ASR.
wav2vec 2.0 \cite{baevski2020wav2vec} proposes using contrastive learning to predict representations of masked parts.
This method, while proven effective across diverse domains \cite{babu2022xls,huang2023wav2vec}, is computationally expensive.
Instead of using contrastive learning, HuBERT \cite{hsu2021hubert} employs k-means clustering to learn a quantizer that maps speech signals into discrete labels. 
It then performs BERT-style \cite{devlin2019bert} prediction pre-training, by masking regions of the input.
The prediction loss is applied over the masked regions.
%Variants of HuBERT have emerged as to simplify its pre-training \cite{lin2023melhubert} or make it speaker-aware \cite{chen2022unispeech}. 
Both wav2vec 2.0 and HuBERT have been utilized to improve speech recognition for low-resource domains \cite{zhao2022improving}.
%HuBERT \cite{hsu2021hubert} employs k-means clustering to provide aligned target labels for pre-training of a BERT model. 
%The predictive loss over the masked regions only is applied \cite{devlin2019bertpretrainingdeepbidirectional}.
The use of semi-supervised with self-supervised training has been explored for unseen domain adaptation \cite{hwang22c_interspeech,hwang2022large}.
More recently, BERT-based Speech pre-Training with Random-projection Quantizer (BEST-RQ) \cite{chiu2022self} was introduced as a lighter SSL alternative for training speech encoders. 
The approach learns a model to predict the masked speech signals, in the form of discrete labels generated with a random-projection quantizer. The quantizer projects speech inputs with a randomly initialized matrix, and does a nearest-neighbor lookup in a randomly initialized codebook.
As the quantizer is kept frozen and independent from the ASR model, this strategy remains adaptable and suitable for integration with any speech recognition framework.
BEST-RQ has showed speech recognition improvements in different domains \cite{zhang2023google,huang2025nest}.

%, leaving decoder design for later supervised adaptation.

In the context of a project funded by the French Directorate General of Armaments, we focus on the Air Traffic Control (ATC) communications low-resource domain.
The ATC domain presents unique challenges for ASR: non-native speech, noisy conditions, and specialized phraseology.
%In this domain, pilots and air traffic controllers from all over the world communicate in English to ensure flight safety.
Transcribed data from this domain are scarce, whereas large amounts of untranscribed ATC speech are readily available \cite{zuluaga2022atco2}.
In order to improve ATC speech recognition, different works have been proposed over the years, on simulated data \cite{ferreiros2012speech}, using a complex cascaded architecture \cite{9174746}, through a challenge \cite{pellegrini2019airbus}, and using diverse datasets \cite{badrinath2022automatic}.
%\cite{ferreiros2012speech,9174746, pellegrini2019airbus,badrinath2022automatic}. 
XLS-R and Whisper were studied on ATC data and showed promising results \cite{zuluaga2023does,van2024whisper}. 
All these previous works are based on the use of labeled ATC data only.
%leaving largely available unlabeled data unused.
Seeing how Whisper already provides a strong base for ATC ASR, we propose to leverage a large amount of unlabeled ATC speech to further improve Whisper's speech recognition. 

\begin{figure*}[htb]
    \centering
    \centerline{\includegraphics[scale=0.875]{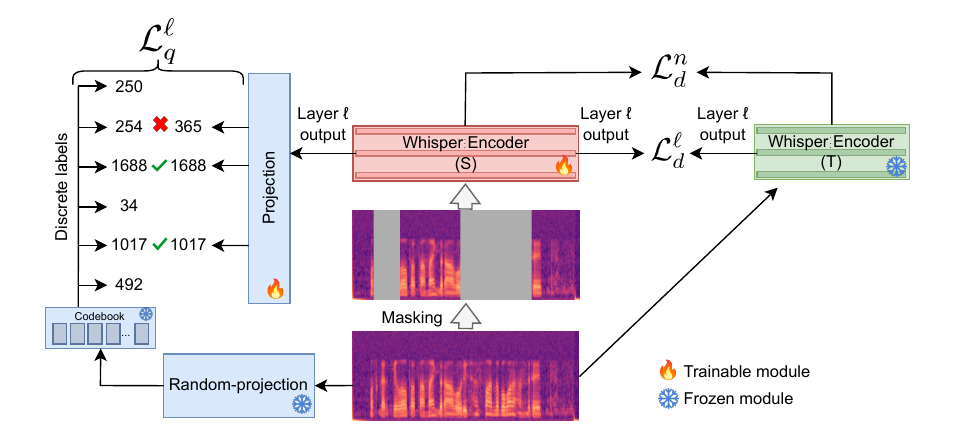}}
    \caption{Architecture of the proposed BEARD framework. On the left side, we use BEST-RQ's objective ($\mathcal{L}_q^\ell$). It is applied to the output of the $\ell$-th Transformer layer. On the right, we use two distillation losses: $\mathcal{L}_d^\ell$, $\mathcal{L}_d^n$. They are computed at two different layers, the $\ell$-th layer and the output layer, respectively, by leveraging a frozen teacher encoder.}
     \label{fig:archi}
\end{figure*}

In this paper, we propose BEARD (BEST-RQ Encoder Adaptation with Re-training and Distillation), a framework that adapts Whisper’s encoder with unlabeled speech to improve ASR performance on new domains.
Unlike traditional SSL methods, which are designed to pre-train speech encoders from scratch, BEARD makes use of Whisper’s already powerful encoder.
By combining self-supervised learning and distillation, BEARD re-trains Whisper's encoder on untranscribed speech while preserving its complementarity with the decoder.
We show that BEARD gives a stronger base to fine-tune on low-resource labeled data from the same domain.
Using BEARD decreases significantly the word error rate (WER) on the ATCO2 corpus w.r.t.\ previous works \cite{zuluaga2023does,van2024whisper} and a model fine-tuning using only labeled data.
W.r.t.\ previous studies, we apply an SSL objective to a middle-layer of the speech encoder and use distillation at the same time.
To the best of our knowledge, this is the first work that leverages self-supervised learning to improve Whisper.

%The paper is organized as follows. Section~\ref{sec:proposed_method} introduces the proposed method. Section~\ref{sec:exp_settings} details the experiments. Section~\ref{sec:results_discu} presents our results. We conclude in Section~\ref{sec:conclusion}.

\section{Proposed methodology}
\label{sec:proposed_method}

\begin{table*}[th]
  \caption{WER (\%) obtained with different fine-tuning methods on ATCO2. The Adaptation method column indicates the method used to adapt the ASR model. The ATCO2 for FT column indicates the quantity of ATCO2 audio used for fine-tuning.
  For BEARD, we indicate how many hours of unlabeled audio were used in parentheses.
  Bold numbers indicate the best result and those results which are statistically equivalent to it according to SCTK's matched pair sentence segment test. * indicates results reported in the literature.}
  \label{tab:results_big}
  \centering
  \begin{tabular}{ c c c c c c }
    \toprule
    {\textbf{Adaptation method}} & \textbf{\# FT params.} & \textbf{ATCO2 for FT} & \textbf{Layer $\ell$} & \textbf{Distillation weight $\bm\lambda$} & {\textbf{WER (\%)}} \\
    \midrule
    \midrule
    XLS-R (FT ATC) + LM \cite{zuluaga2023does} & 300M & 0 min & & & 19.80* \\
    Whisper-small, FT \cite{van2024whisper} & 244M & 52 min & & & 22.79* \\
    %Whisper-medium-based \textbf{A ENLEVER} \cite{van2024whisper} & 769M & - & - & \textbf{17.99}* \\
    %\midrule
    \midrule
    Whisper-small, no FT & 0 & 0 min & & & 63.32 \\
    Whisper-small, FT & 244M & 2 h 24 min & & & 19.54 \\
    \midrule[1.5pt]
    \multirow{5}{*}{\shortstack{Whisper-small,\\BEARD (5381 h) + FT}} & \multirow{5}{*}{244M} & \multirow{5}{*}{2 h 24 min} & 8 & 0.5 / 1.0 & 18.40 / 18.27 \\
      %\cmidrule(lr){4-6}
      & & & 7 & 0.5 / 1.0 & \textbf{17.87} / 19.68 \\
      %\cmidrule(lr){4-6}
      & & & 6 & 0.5 / 1.0 & \textbf{17.17} / 18.05 \\
      %\cmidrule(lr){4-6}
      & & & 5 & 0.5 / 1.0 & \textbf{17.72} / 18.06 \\
      %\cmidrule(lr){4-6}
      & & & 4 & 0.5 / 1.0 & 18.01 / \textbf{17.58} \\
      
    \midrule
     \makecell{Whisper-small,\\BEARD (2000 h/1000 h/500 h) + FT} & \multirow{0.5}{*}{244M} & 2 h 24 min & 6 & 0.5 & 18.40 / 18.06 / \textbf{17.53} \\
   %\midrule
     %\shortstack{Whisper-small,\\BEARD (1000h) + FT} & 244M & 2 h 24 min & 6 & 0.5 & 18.06 \\
   %\midrule
     %\shortstack{Whisper-small,\\BEARD (500h) + FT} & 244M & 2 h 24 min & 6 & 0.5 & \textbf{17.53} \\
   \bottomrule
  \end{tabular}
\end{table*}

\subsection{Whisper}

Whisper, end-to-end encoder-decoder Transformer, is a state-of-the-art model for automatic speech recognition \cite{radford2023robust}.
It has been trained on 680,000 hours of transcribed multilingual data. 
Whisper's encoder is mostly focused on acoustic features, while its decoder is mostly focused on linguistic features.
Whisper differs from self-supervised learning models in that it is trained exclusively on labeled data, using a supervised objective.

\subsection{BEST-RQ}

BERT-based Speech pre-Training with Random-projection Quantizer (BEST-RQ) \cite{chiu2022self} is a self-supervised learning approach which converts input speech signal into discrete labels.
This is done with a random-projection quantizer, which projects the input signal and finds the nearest vector in a codebook.
The index of that vector becomes the discrete label.
During pre-training, portions of the input log-mel spectrogram are masked and fed into a speech encoder.
The speech encoder is trained to predict the discrete labels of the masked parts.
%It operates by masking portions of the input log-mel spectrogram and training a speech encoder to reconstruct the masked regions from the surrounding context, using labels generated by a random-projection quantizer as learning targets.
The random-projection quantizer is kept frozen throughout the pre-training stage, which makes it easier to train than wav2vec2.0 or HuBERT's.
After pre-training the speech encoder, a decoder can be added on top and trained for the ASR task.
BEST-RQ had first been proposed as a pre-training approach for Conformer architectures \cite{gulati2020conformer}.
It was demonstrated that BEST-RQ can also be applied to convolution-free Transformer architectures \cite{hou2024revisiting}.

\subsection{BEST-RQ Encoder Adaptation with Re-training and Distillation (BEARD)}

%When labeled data is scarce, but unlabeled data is abundant, self-supervised learning (SSL) methods are particularly effective.
SSL approaches, such as BEST-RQ, cannot be directly applied to models like Whisper.
These methods are designed to train speech encoders from scratch, without a pre-existing decoder.
In Whisper's case, the encoder was trained along the decoder.
Applying SSL to the encoder would leave the decoder unchanged, leading to a mismatch between them.

%In Whisper’s case, this would discard a decoder trained on 680,000 hours of transcribed speech.
%Discarding its highly performing decoder would negate much of Whisper’s value (\textbf{reformuler)}.
%Applying BEST-RQ to a model that was originally trained under a completely different paradigm, such as the encoder–decoder supervised ASR model Whisper, is non-trivial.
%As BEST-RQ only applies to speech encoders, it leaves behind the decoder and guarantees no compatibility with it afterwards.
%We would then have to add a new decoder on top and train it again, completely discarding previous decoding ability.
%With models like Whisper, trained on 680,000 hours of transcribed data, this is a huge loss.
%Directly applying BEST-RQ to the speech encoder and subsequently fine-tuning both encoder and decoder is risky, as the encoder’s representations may diverge from what the decoder expects, leading to degraded recognition.

To address this issue, we propose BEARD, a framework that leverages BEST-RQ self-supervision to adapt Whisper’s encoder with untranscribed data, combined with knowledge distillation to maintain complementarity with its decoder.
As shown in Fig.~\ref{fig:archi}, we take two copies of Whisper's pre-trained encoder, the decoder is not used.
One of the encoders (referred to as S, for student) will be modified using unlabeled data, we call this entire stage \textit{re-training}. 
The other encoder (referred to as T, for teacher) is frozen and will serve as the teacher for the distillation.

To re-train S with unlabeled data, we apply BEST-RQ's quantization mechanism, whose loss is noted as $\mathcal{L}_q^\ell$.
Following Chiu et al. \cite{chiu2022self}, a frozen random-projection quantizer is used, and by using codebooks, assigns discrete labels to the input.
Instead of applying the prediction loss to the output of S, as originally proposed \cite{chiu2022self}, we propose applying it to the output of its $\ell$-th layer, out of $n$ layers.
This allows the upper layers to maintain representational alignment with the decoder, preserving the complementarity.
We add a projection layer that takes the output of the $\ell$-th layer as input, and is trained to predict the discrete labels of masked regions, as in BEST-RQ.

%Specifically, BEARD applies a self-supervised random-projection quantizer loss to encourage representation learning from unlabeled speech, while simultaneously leveraging distillation to preserve the knowledge of the original model and guide the adaptation process.
%As shown in Fig.~\ref{fig:archi}, we propose applying a BEST-RQ quantization loss $\mathcal{L}_q^\ell$ to an intermediate encoder layer $\ell$, with $\ell \in \{0,...,n-1\}$ out of $n$ encoder layers (12 for Whisper-small), thereby injecting knowledge from unlabeled data. 

Using only $\mathcal{L}_q^\ell$ would lead to changes in the model up to the $\ell$-th layer, leaving the following layers untouched.
In order to adapt all layers, and, at the same time, preserve the complementarity with the decoder, we introduce two distillation losses using the frozen teacher encoder (T).
%As we want to keep compatibility with the decoder, 
$\mathcal{L}_d^n$ is computed between the last layer outputs of S and T.
By making sure S's output space remains close to T's, we assure the complementarity with the decoder is preserved.
$\mathcal{L}_d^\ell$ is computed between the outputs of the $\ell$-th layer of S and the $\ell$-th layer of T, which helps the distillation process.
%This is discussed in Section~\ref{sec:ablation}.
%To prevent the encoder from drifting too far from the representations expected by the decoder, we introduce two distillation losses using the original frozen encoder: $\mathcal{L}_d^\ell$, applied to the hidden representations at layer $\ell$ of the training and frozen encoders, and $\mathcal{L}_d^n$, applied at the encoder output layer.
Both $\mathcal{L}_d^n$ and $\mathcal{L}_d^\ell$ use cosine similarity, which should be maximized.
We use cosine similarity over L1 or mean squared error, because it is less constraining as it does not penalize vector norms, allowing for the encoder to adapt to the new domain.
Furthermore, as using SSL requires masking parts of S's input, the distillation losses are computed exclusively on the unmasked regions.
%Furthermore, these two losses were applied exclusively to the unmasked regions of the input.
%This is because they are not made to reconstruct the masked parts like $\mathcal{L}_q^\ell$, but to make sure the existing projections stay close to the teacher's.

The overall training objective $\mathcal{L}$ is defined as
\begin{align}
    \mathcal{L} &= \mathcal{L}_q^\ell + \lambda \mathcal{L}_d^\ell + \beta \lambda \mathcal{L}_d^n
\label{eq:train_loss}
\end{align}
where $\lambda$ and $\beta$ are weighting coefficients.
%We use a weaker penalty on the output layer ($\lambda/10$) to gently anchor the decoder interface. 
%This prevents excessive drift of the encoder’s outputs toward unfamiliar states, yet avoids over-regularizing the adaptation driven by the quantizer.

Following Chiu et al. \cite{chiu2022self}, we apply normalization to prevent the random projection from collapsing to a limited subset of codes.
The normalization is applied using LayerNorm at the input of both the random-projection quantizer and the projection layer.
It normalizes the vectors to have 0 mean and standard deviation of 1.
%Following the original BEST-RQ implementation, and to prevent the random projection from collapsing to a limited subset of codes, we inserted a LayerNorm both at the input of the random-projection quantizer and at the input of the projection layer (see Fig.~\ref{fig:archi}).
%The distillation losses were applied exclusively to the unmasked regions of the input, ensuring that the model was not penalized for errors on spans where the input had been corrupted.

After re-training S using BEARD, we propose adding back Whisper's decoder on top of S.
We then fine-tune them both, without BEARD, using labeled speech from the same domain as the untranscribed speech.
Compared to previous works, we adapt a pre-trained supervised model. 
\cite{hwang22c_interspeech} proposes joint SSL/supervised training, while \cite{hwang2022large} explores SSL+FT sequentially. 
In our approach, we use SSL to adapt a pre-trained supervised model along with distillation.
\section{Experimental settings}
\label{sec:exp_settings}
\textbf{Dataset ---}
%\subsection{Dataset}
We conducted our experiments using the ATCO2 dataset\footnote{ATCO2 project data, ELRA catalog (\url{http://catalog.elra.info}), ISLRN : 589-403-577-685-7, ELRA ID : ELRA-S0484} \cite{zuluaga2022atco2}. 
It contains air traffic control communications between pilots and air traffic controllers from various airports.
% (Australia, Czech Republic, Slovakia, and Switzerland).
The speech is non-native, with high speech rate and noisy, with signal-to-noise ratios (SNR) varying from -10~dB to 40~dB, estimated using WADA-SNR \cite{kim2008robust}.
The corpus consists of 2 parts:
%the first part, ...
5,381 hours of audio without transcription,
%the second part
and 4 hours of audio with human transcriptions\footnote{Only one hour of transcribed audio is freely available.}.
We do not have any information about the number of speakers and their native language.
For the self-supervised re-training stage, we employ the untranscribed part of the dataset.
For the fine-tuning stage, we use the transcribed speech part.
%Due to the small amount of transcribed data available, 
We conduct 4-fold cross-validation.
For a given fold, 2 h 24 min (25,000 words) are used for the fine-tuning, 36 min (5,300 words) for validation, and 1 h (10,000~words) for testing.
All the audio files of ATCO2 are sampled at 16~kHz, which matches Whisper's input requirements.
%To ensure ASR evaluation across all available transcribed data, we conducted 4-fold cross-validation, with each fold’s test set containing 1 hour of audio.
%The 3 hours left within each fold were used for fine-tuning and validation, using a 80/20 ratio.
%ajouter nombre de mots

%Since some audio files in the dataset include multiple speaking turns, we segmented them into smaller files, each containing a single turn. 
%To preserve data consistency, all segments originating from the same original file were assigned to the same data split.

%For the self-supervised training stage, we employed the full ATCO2 dataset, comprising 5,381 hours of audio. 
%From this corpus, 5 hours were randomly sampled and set aside for validation.
%For the fine-tuning stage, we utilized the entire transcribed subset of ATCO2, totaling 4 hours of audio. 
%The remaining data within each fold was split into training and validation subsets using an 80/20 ratio.
%Since some audio files in the dataset include multiple speaking turns, we segmented them into smaller files, each containing a single turn. 
%To preserve data consistency, all segments originating from the same original file were assigned to the same data split.
%All the audio files are sampled at 16kHz which matches Whisper's input requirements.

\textbf{General parameters ---}
%\subsection{General parameters}
Our experiments were conducted with the Whisper-small model \cite{radford2023robust} comprising 244M parameters.
Its compact size makes it suitable for deployment on a wide range of devices, including embedded systems, while remaining a high-performance ASR model.
%The pretrained model is publicly available on Hugging Face\footnote{\url{https://huggingface.co/openai/whisper-small}}.
%on le choisit parce qu'on peut le tourner sur différent devices
%cette taille est comparable à la baseline XLS-R
Its parameter scale (244M) is comparable to the XLS-R model \cite{babu2022xls} (300M) employed as the baseline for ATCO2 \cite{zuluaga2023does}.
Moreover, both Whisper-small and XLS-R are end-to-end ASR systems trained on a wide variety of languages.
For BEARD re-training stage, models are re-trained for a single epoch over all untranscribed data available, with a batch size of 32.
We chose to run only one epoch since over 5,000~hours is a significant amount of data.
We set the learning rate to 1e-5 for Whisper's encoder and 5e-4 for the projection layer.
We use a masking span of 4 frames \cite{whetten2024open}.
We reduce the masking probability from 0.15 to 0.10 to ensure that a sufficient number of unmasked frames remain for the distillation.
For the random-projection quantizer, we use a codebook size of 2048, which obtained the best results in our preliminary experiments.
%We tried using a size of 4096, but it showed degraded performance.
We set $\beta$ to 0.1 to down-weight $\mathcal{L}_d^n$.
Higher values led to worse results.
%The explanation could be that the encoder alignment should not be overly constraining. 
%ensuring that the model is guided by the output alignment without being overly constrained
On 8 NVIDIA V100 GPUs, one epoch of BEARD re-training takes around 7 hours on the 5,381~hours of untranscribed speech.
This is much shorter than creating pseudo-labels.
For the fine-tuning stage, models are trained until convergence, with a batch size of 16.
The learning rate is set to 1e-5.
For decoding, we use greedy search for computational reasons.
Our code is publicly available \footnote{\url{https://gitlab.inria.fr/rbagat/beard}}.

%\paragraph{Evaluation metrics}
The results are reported in terms of the WER.
Early stopping is made using the WER on the validation set.
The statistical significance of the results has been validated using the matched pair sentence segment test with SCTK \cite{sctk} ($p=0.001$).

\section{Results and discussions}
\label{sec:results_discu}

\begin{table}[th]
  \caption{WER (\%) on ATCO2 when BEARD is applied without distillation losses, with $\mathcal{L}_d^\ell$ only, with $\mathcal{L}_d^n$ only, or with both. In any case, the $\mathcal{L}_q^\ell$ loss is used. Bold numbers indicate the best result.}
  \label{tab:no_dist}
  \centering
  \begin{tabular}{ c c c c }
    \toprule
    {\textbf{Layer $\bm\ell$ / $\bm\lambda$}} & {\textbf{Using $\mathcal{L}_d^\ell$}} & \textbf{Using $\mathcal{L}_d^n$} & {\textbf{WER (\%)}} \\
    \midrule
    \midrule
    \multirow{4}{*}{6 / 0.5} & No & No & 80.98 \\
     & Yes & No & 37.28 \\
     & No & Yes & 20.44 \\
     & Yes & Yes & \textbf{17.17} \\
    \bottomrule
  \end{tabular}
\end{table}

\subsection{Baselines}

We consider four baselines.
The first two are prior works on the ATCO2 dataset.
An XLS-R model fine-tuned on 132 hours of ATC speech from diverse corpora (referred to as XLS-R FT) \cite{zuluaga2023does}, ATCO2 was only used for testing.
This is the first baseline that was presented on ATCO2, the authors employed an external language model (LM) during decoding.
Van Doorn et al.\ used Whisper-small and fine-tuned it on 52 min of ATCO2 audio (Whisper-small FT) \cite{van2024whisper}.
We evaluate two baselines on Whisper-small: the model without fine-tuning (No FT), and a fine-tuned version (FT) fine-tuned on 2 h 24 min of ATCO2 audio and evaluated with our cross-validation setup.

The reported four baselines are shown in the first four rows of Table~\ref{tab:results_big}. 
Among these, No FT performs worst, with a WER of 63.32\%.
XLS-R FT reaches 19.80\%, outperforming Whisper-small FT (22.79\%). 
FT (2 h 24 min), using the same base model as Whisper-small FT, but fine-tuned with more data, achieves a WER of 19.54\%, which is comparable to XLS-R FT.
As a side experiment, we fine-tuned Whisper-small using pseudo-labels \cite{kahn2020self} from 3,500 hours of untranscribed ATCO2 audio and obtained 25.97\%  WER which is significantly worse than Whisper-small FT (22.79\%).
%Finally, Whisper-medium-based shows the best performance among the baselines, with a WER of 17.99\%, benefiting from 2.5× more parameters.

%They report WERs of 19.80\% and 22.79\%, respectively.
%As a third baseline, we include our own fully fine-tuned Whisper-small (referred to as Full FT) by using the 4 hours of transcribed data.
%By following our cross-validation setup, it obtains a WER of 19.54\%.
%Notably, Full FT reaches the best baseline performance, improving upon \cite{van2024whisper} by leveraging more data.

\subsection{BEARD results}
\label{sec:beard_res}

%\subsubsection{Effect of $\ell$}
For every experiment, we take the pre-trained Whisper-small model and we re-train its 12-layer encoder using BEARD.
Then, we fine-tune both encoder and decoder on transcribed speech in a cross-validation setup.
We evaluated BEARD, by using 5,381 h of unlabeled audio, under multiple conditions with different parameter settings: encoder layer $\ell \in \{4,5,6,7,8\}$ and weighting factor $\lambda \in \{0.5, 1.0\}$.
These values were chosen to evaluate the sensibility of the method to the choice of the layer $\ell$, and to see whether the distillation should be more balanced or not.

%For every experiment, after adapting Whisper’s encoder with BEARD, we reattached the decoder and fine-tuned the entire model using transcribed data in a cross-validation setup.

\textbf{Effect of $\bm\ell$} ---
Results in Table~\ref{tab:results_big} show that applying BEARD at layers $\ell \in \{4,5,6,8\}$ and using $\lambda=0.5$ consistently significantly outperforms baseline results.
Our best configuration, with $\ell=6$ and $\lambda=0.5$, achieves a WER of \textbf{17.17\%}, which represents a \emph{significant improvement} of 12\% (relative) compared to FT (19.54\%).
%This is a 12\% relative improvement compared to Full FT.
%matching the performance of Whisper-medium-based despite requiring substantially fewer parameters.
Applying BEARD at layers 4, 5, and 6 yields slightly better results than at layer 7 and 8, suggesting that middle-level encoder layers are more effective for this type of adaptation.
In additional experiments not reported in the table, applying BEARD at layer $\ell = 9$ with $\lambda=0.5$, led to a WER of 18.64\%.
This further confirms that performance tends to degrade as BEARD is applied closer to the encoder's output.

%\subsubsection{Effect of $\lambda$}

\textbf{Effect of $\bm\lambda$} ---
To assess the influence of the distillation weight $\lambda$, we explored different values. We only report $\lambda=0.5$, which is the best value we found, and $\lambda=1.0$ for comparison, for each of the encoder layer $\ell$ to which BEARD was applied.
%we evaluated two values, $\lambda \in \{0.5, 1.0\}$, for each of the encoder layer $\ell$ to which BEARD was applied.
As shown in Table~\ref{tab:results_big}, lower value of $\lambda$ leads to better performance across most layers. 
For layers $\ell = 5$, $6$, and $7$, using $\lambda = 0.5$ resulted in significantly lower WERs compared to $\lambda = 1.0$.
%Interestingly, the trend does not hold at layer $\ell = 8$ and 4, where both values of $\lambda$ yielded similar performance (18.40\% versus 18.27\%, and 18.01\% versus 17.58\%).
In all cases, except when BEARD is applied at layer 7 with $\lambda = 1.0$, the results largely outperform FT.

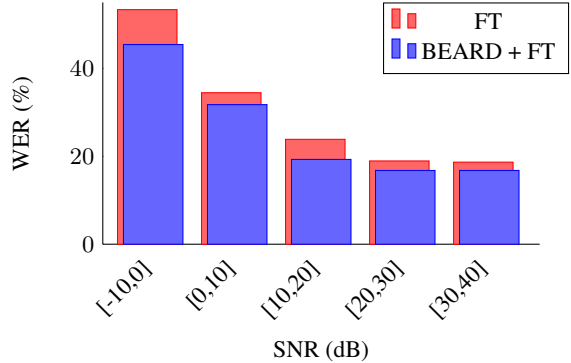
\begin{figure}[ht]
\centering
\begin{tikzpicture}
\begin{axis}[
    ybar,
    bar width=15pt,
    scale=0.5, %0.565
    xscale=1.4, %1.5
    enlarge x limits=0.15,
    ymin=0,
    ymax=55,
    ylabel={WER (\%)},
    xlabel={SNR (dB)},
    symbolic x coords={{[-10,0]}, {[0,10]}, {[10,20]}, {[20,30]}, {[30,40]}},
    xtick=data,
    x tick label style={
        rotate=45,
        anchor=east,
        yshift=-4pt, % -4pt
        xshift=-4pt % -4 pt
    },
    legend style={
        at={(0.7,1)},         % top right of the axis
        anchor=north east,  % align top-right corner of legend box to that point
        %font=\scriptsize,
        %draw=none           % optional: remove legend box border
    },
    %nodes near coords,
    %nodes near coords align={vertical},
    axis y line*=left,
    axis x line*=bottom,
    every node near coord/.append style={font=\small},
    major tick length=0pt
]

\addplot+[red,fill=red!60,bar shift=-1.5pt]
    coordinates {({[-10,0]},53.36) ({[0,10]},34.46) ({[10,20]},23.86)
        ({[20,30]},18.94) ({[30,40]},18.67)};
\addlegendentry{FT}

\addplot+[style={blue,fill=blue!60,bar shift=0pt}] 
    coordinates {({[-10,0]},45.43) ({[0,10]},31.75) ({[10,20]},19.29) ({[20,30]},16.78)
                 ({[30,40]},16.78)};
\addlegendentry{BEARD + FT}

\end{axis}
\end{tikzpicture}
\caption{Comparison of WER across SNR bins for our best BEARD configuration ($\ell=6,\lambda=0.5$) and FT. SNR was estimated using WADA-SNR \cite{kim2008robust}.}
\label{fig:snr}
\end{figure}

\textbf{SNR analysis} ---
As previously mentioned, one of the main challenges of the ATC domain is the noise.
Our ATCO2 unlabeled data contains different SNR level indications.  
We expect the proposed methodology will perform better across different SNRs. 
Fig.~\ref{fig:snr} compares the WER obtained by our best BEARD configuration ($\ell=~6, \lambda=0.5$) and the FT model at different SNR levels.
The SNRs were computed using WADA-SNR \cite{kim2008robust}.
%We report results on SNRs varying from -10dB to 40dB because the lower/higher values do not contain sufficient samples for results to be meaningful.
As it can be seen, the model re-trained with BEARD outperforms FT across all SNR levels.
The largest improvements are within the [10,20]dB SNR range, with a relative improvement of 19\%.
Notably, in the negative SNRs, BEARD reaches a relative improvement of 15\% over FT.

%snr_bin
% BEARD
%(-10, 0]      0.454284
%(0, 10]       0.317537
%(10, 20]      0.192853
%(20, 30]      0.163869
%(30, 40]      0.167819
%(40, 50]      0.171741
%(50, 60]      0.089312

%snr_bin
% ORIGINAL FULL-FT
%(-10, 0]      0.533586
%(0, 10]       0.344647
%(10, 20]      0.238563
%(20, 30]      0.189405
%(30, 40]      0.186673
%(40, 50]      0.228125
%(50, 60]      0.170360

%\subsection{Ablation study}

\textbf{Effect of data quantity for BEARD ---} We also applied BEARD using different amounts of unlabeled data in our best configuration, as reported in the last row of Table \ref{tab:results_big}, and obtained WERs of 18.40\%, 18.06\% and 17.53\%.
This shows that BEARD is also beneficial when less unlabeled data are available.

\textbf{Ablation study} ---
\label{sec:ablation}
We conducted an ablation study to assess the contribution of the distillation losses. 
Four variants were compared: a version of BEARD without any distillation, one using only $\mathcal{L}_d^\ell$,  one using only $\mathcal{L}_d^n$, and the proposed BEARD combining both losses. 
For all variants, the $\mathcal{L}_q^\ell$ loss was used during re-training on unlabeled data, and the models were fine-tuned on labeled data after.
All experiments were performed by applying BEARD at layer $\ell=6$ with $\lambda=0.5$, which is our best configuration.
%Re-trained models using BEARD are then fine-tuned with our cross-validation setup.
Results can be found in Table~\ref{tab:no_dist}.
It can be seen that removing distillation entirely leads to heavy degradation, with a WER of 80.98\%.
It confirms that BEST-RQ's loss should not be applied to Whisper's encoder without assuring complementarity with the decoder.
Using distillation with only $\mathcal{L}_d^\ell$ , still results in degraded performance, with a WER of 37.28\%.
By using only $\mathcal{L}_d^n$, the results improve over using only $\mathcal{L}_d^\ell$ and lead to a WER of 20.44\%,
which is comparable to the FT model.
This exhibits that when the complementarity with the decoder is not sufficiently assured, applying BEARD has less effect on the ASR performance.
The best result is obtained when using both distillation losses, reaching 17.17\% WER.
This shows the importance of having distillation within the encoder, along the self-supervised objective.

\section{Conclusion}
\label{sec:conclusion}

We investigated whether self-supervised learning can help Whisper adapt to a new domain.
We introduced BEARD, a framework that combines self-supervised learning and distillation to adapt Whisper's encoder using unlabeled speech.
The modified encoder is then fine-tuned with the decoder using a limited amount of labeled data.
%Our experiments showed that BEARD significantly improves over both baselines and full fine-tuning. 
On the ATCO2 corpus, the best BEARD configuration achieved a relative 12\% WER improvement over fine-tuning with transcribed speech only. 
BEARD showed WER improvements across all SNRs.
The ablation study demonstrated that distillation is important to keep the encoder-decoder complementarity.
These results confirm that large-scale unlabeled data can be effectively exploited through self-supervision to adapt Whisper ASR model.
%such as Whisper.
The proposed approach can be used to adapt any encoder-decoder models to new domains. 
%Future work may explore extending this approach to other domains where labeled data is scarce.
Future work will study other SSL objectives, multi-layer SSL adaptation,
%use different quantity of unlabeled data (not currently tested due to computational limitations), 
filter the unlabeled data, and adapt a larger Whisper.
%may explore scaling our approach to larger Whisper variants, testing alternative SSL objectives, and extending to multilingual data where annotations can be rare.
%On l'applique sur ATC Whisper, mais devrait fonctionner sur autres domaines et modèles encoder/decoder pré-entrainé

%\section{Acknowledgments}

%This work was funded by the DeepMAUVES project supported by DGA of french MoD and CNRS, and granted access to the HPC resources of IDRIS under the allocation 2024-AD011015024 made by GENCI.

%\vfill\pagebreak

%\section{Acknowledgments}
%This work was funded by the DeepMAUVES project supported by DGA of French MoD and CNRS, and granted access to the HPC resources of IDRIS under the allocation 2024-AD011015024 made by GENCI.

%\section{Acknowledgments}
%This work was granted access to the HPC resources of IDRIS under the allocation 2024-AD011015024 made by GENCI.

\bibliographystyle{IEEEbib}
\bibliography{strings,refs}

\end{document}